\documentclass{article}

\usepackage{arxiv}

\usepackage[utf8]{inputenc} 
\usepackage[T1]{fontenc}    
\usepackage{hyperref}       
\usepackage{url}            
\usepackage{booktabs}       
\usepackage{amsfonts}       
\usepackage{nicefrac}       
\usepackage{microtype}      
\usepackage{lipsum}
\usepackage{graphicx}
\usepackage{multirow}
\usepackage{amsmath}
\usepackage{xcolor}
\graphicspath{ {./images/} }

\title{Contactless estimation of continuum displacement and mechanical compressibility from image series using a deep learning based framework}

\author{
 Anto Nivin Maria Antony\\
  Leibniz Institute for Plant Genetics and Crop Plant Research (IPK)\\
  Seeland, Germany\\
  \texttt{maria@ipk-gatersleben.de} \\
   \And
 Thomas Richter\\
  Otto-von-Guericke Universität\\
  Magdeburg, Germany\\
  \texttt{thomas.richter@ovgu.de}\\
  \And
 Evgeny Gladilin\\
  Leibniz Institute for Plant Genetics and Crop Plant Research (IPK)\\
  Seeland, Germany\\
  \texttt{gladilin@ipk-gatersleben.de} \\
}

\begin{document}
\maketitle
\begin{abstract}
Contactless and non-invasive estimation of mechanical properties of physical media from optical observations is of interest for manifold engineering and biomedical applications, where direct physical measurements are not possible. 
Conventional approaches to the assessment of image displacement and non-contact material probing typically rely on time-consuming iterative algorithms for non-rigid image registration and constitutive modelling using discretization and iterative numerical solving techniques, such as Finite Element Method (FEM) and Finite Difference Method (FDM), which are not suitable for high-throughput data processing. Here, we present an efficient deep learning based end-to-end approach for the estimation of continuum displacement and material compressibility directly from the image series. Based on two deep neural networks for image registration and material compressibility estimation, this framework outperforms conventional approaches in terms of efficiency and accuracy. In particular, our experimental results show that the deep learning model trained on a set of reference data can accurately determine the material compressibility even in the presence of substantial local deviations of the mapping predicted by image registration from the reference displacement field. 
Our findings suggest that the remarkable accuracy of the deep learning end-to-end model originates from its ability to assess higher-order cognitive features, such as the vorticity of the vector field, rather than conventional local features of the image displacement.  
\end{abstract}

\section{Introduction}
Contactless analysis of mechanical structures, such as engineering materials and biological tissues, is a method of choice when direct physical probing is not possible or applicable \cite{epub90639, mattana2018non}. This is, in particular, the case when target structures can only be monitored remotely using non-invasive optical observation techniques such as magnetic resonance elastography \cite{muthupillai1996magnetic}, ultrasound elastography \cite{ophir1999elastography}, ultrasound shear wave elastography \cite{sarvazyan1998shear}, etc. Non-contacting optical monitoring enables significantly higher throughput of measurements compared to contacting techniques; however, it provides indirect observations in the form of image sequences and requires additional image analysis and numerical simulation to reveal hidden parameters of underlying constitutive models.

In a number of previous works, including the author's early findings \cite{gladilin20073d, gladilin2010contactless, fougeron2024current}, approaches to image- and model-based estimation of material properties of optically monitored structures was reported. These approaches are typically based on image registration followed by fitting a numerical model of material behaviour to a pair of co-registered images. Depending on the particular data modality and task, whole image or boundary (contour re. surface) based non-rigid registration in combination with conventional iterative solving techniques such as Finite Difference or Finite Element Method is applied. The pipelines of conventional image analysis and numerical problem solving are, however, known to be time-consuming and cumbersome (especially with respect to individual mesh generation), which substantially limits their high-throughput capabilities \cite{yue2013tissue, roters2010overview, diehl2017identifying}. In contrast, modern screening methods generate a large amount of real-time images that require efficient high-throughput algorithmic solutions. 

The understanding of the material properties of cells is the fundamental basis in the fields of cell mechanics and mechanobiology \cite{javanmardi2021quantifying}. The biological behaviours of the cells, like division, deformation, and differentiation, are related to the material properties of cells \cite{lange2013cell, huang2004cell}. The material property is mostly explained with the help of Young's modulus and Poisson's ratio \cite{javanmardi2021quantifying, islam2019estimation, islam2020non}. The stiffness is one parameter that is extensively studied, given by Young's modulus\cite{luo2016cell}. However, the Poisson's ratio which explains the material compressibility is mostly overlooked \cite{javanmardi2021quantifying}. The estimation of Poisson's ratio is crucial for cancer diagnosis, prognosis, and treatment as explained in \cite{islam2019estimation, islam2020non}.

In the past decades, numerous image registration methods for uni- and multi-model rigid and non-rigid alignment were developed \cite{Zitova03, xu2018review, dong2021review, wang2020non}. Conventional methods of image registration are computationally expensive \cite{kanter2022flexible}, in particular, when applied to images of high resolution when convergence of the iterative optimization solvers becomes slow. Recent advances in machine learning, in particular, deep learning, open new opportunities for efficient computational solutions. Deep learning has been used for efficient implementation of 2D and 3D image registration \cite{de2017end, balakrishnan2019voxelmorph}. However, investigations of the accuracy of deep learning registration methods in application to concrete registration problems are rarely performed. 

The application of deep learning to continuum and computational mechanics has been comprehensively reviewed in \cite{bock2019review, herrmann2024deep}. In \cite{yang2020prediction}, the stress-strain patterns were studied with Convolutional Neural Networks (CNN) along with principal component analysis, and in \cite{rashid2022learning} it is studied in digital composites using Fourier Neural Operator (FNO) in an end to end fashion. The studies were performed in a supervised fashion with a pair of input geometries and ground truth solutions. To our best knowledge, determining Poisson's ratio from images hasn't been studied yet.


In this study, we present a framework for fully automated estimation of material compressibility of optically monitored structures which is based on a combination of deep learning non-rigid image registration trained on a pair of source target-images and deep learned-based regression model (\cite{RFB15a}) trained to assign 2D whole displacement fields generated from Finite Difference scheme to one of the three material compressibilities corresponding to low, high and intermediate compressibility. Throughout the study, we assume the material medium is isotropic and homogeneous.

The manuscript is organized as follows: first, we describe the basic methods of this study, including the linear elastic Finite Difference Method, deep learning based non-rigid image registration scheme, as well as the deep learning-based model for material compressibility estimation. Then, the results of our investigations are based on the reconstruction of the displacement field from images and, the estimation of material compressibility followed by discussion.    


\section{Methods}
This section presents the methods we used for
generation of reference data (i.e. linear elastic displacement fields, deformed images), deep learning based
non-rigid image registration, and estimation of Poisson's ratio.
\subsection{Linear elastic deformation model - 2D Lam{\'e}-Navier PDE}
In the case of a one-material problem and boundary forces that are given implicitly as prescribed boundary displacements
(i.e. $\textbf{f}=0$ and $\textbf{u}=\textbf{u}_{\text{known}}+\textbf{u}_{\text{unknown}}$),
Eq. \ref{eq:lame-navier-pde-geo} takes a quasi-geometric form
\begin{equation}
\label{eq:lame-navier-pde-geo}
\,-(1-2\nu)\, \boldsymbol{\Delta}{u} - \text{grad} \, \text{div} \, {u} = 0\,,
\end{equation}   
with the displacement ${\bf u}$ as  a single variable and the Poisson's ratio $\nu$ as single material parameter.
Eq. \ref{eq:lame-navier-pde-geo} with arbitrary boundary conditions is solved using the Finite Difference Method on the regular 2D image grid. In the case of a quasi-geometrical (pure displacement), the Finite Difference scheme of the Lam{\'e}-Navier PDE takes the following form
%
%
%
%
       \begin{equation}
       \label{eq:fdmscheme}
       \begin{gathered}
       2(1-\nu)\left[\frac{u_x(i+1,j) -2u_x(i,j)+u_x(i-1, j)}{\Delta x^2}\right]\ +\ (1-2\nu)\left[\frac{u_x(i,j+1) -2u_x(i,j) + u_x(i, j-1)}{\Delta y^2}\right] \\
         +\ \left[\frac{u_y(i+1,j+1) -u_y(i+1,j-1) - u_y(i-1, j+1) + u_y(i-1, j-1)}{4 \Delta x \Delta y}\right]\ = 0\,,
        \\
       2(1-\nu)\left[\frac{u_y(i,j+1) -2u_y(i,j)+u_y(i, j-1)}{\Delta y^2}\right]\ +\ (1-2\nu)\left[\frac{u_y(i+1,j) -2u_y(i,j) + u_y(i-1, j)}{\Delta x^2}\right] \\
       +\ \left[\frac{u_x(i+1,j+1) -u_x(i+1,j-1) - u_x(i-1, j+1) + u_x(i-1, j-1)}{4 \Delta x \Delta y}\right]\ = 0\,.
      \end{gathered}
      \end{equation}
 The FDM scheme transforms the continuous second-order 2D Lam{\'e}-Navier PDE to the linear system of equations with respect to unknown values of xy-components of the displacement:
%
       \begin{equation}
        A\,{u} = {b}\,, 
      \label{eq:matform}
      \end{equation}
 where $A$ is a sparse symmetric and positive definite matrix (the so-called stiffness matrix), 
 ${b}$ is the right-hand side vector resulting from the implementation of known boundary conditions, and
 ${u}$ is the vector of unknown displacements.
 In the case of a single-material problem the  Eq. \ref{eq:lame-navier-pde-geo} contains only one material parameter, - the Poisson's ratio ($\nu \in [0,0.5]$)
 describing the material compressibility. In this work, three distinctive values of $\nu = 0, 0.25, 0.49$  were chosen for the simulation of high, moderate, and low compressible material behaviour, respectively.


%


\subsection{Generation of reference source - target deformed images}
The displacement fields obtained using the Finite Difference Method for arbitrary boundary value problems are used for the generation of pairs of source $S$ (undeformed) and target $T$ (deformed) images, where the target images represent the deformed configuration of source images after application of the displacement (${u}$) computed using the FDM, i.e.
      \begin{equation}
       T({x}) = S({x}+{u}). 
      \label{eq:source_target_img}
      \end{equation}
The non-rigidly deformed target images were computed from source images as described in Eq. \ref{eq:source_target_img} using the MATLAB \textit{imwarp} function, i.e. $T$=\textit{imwarp}($S$,{u}). 

\subsection{Unsupervised image registration using CNN}
The deep learning image registration framework used in this work was derived from Voxelmorph \cite{balakrishnan2019voxelmorph}, 
which takes a pair of source and target images, and predicts the displacement mapping the source onto the target image as well as the registered image.
\begin{figure}[ht!]
\begin{center}
\includegraphics[width=18cm]{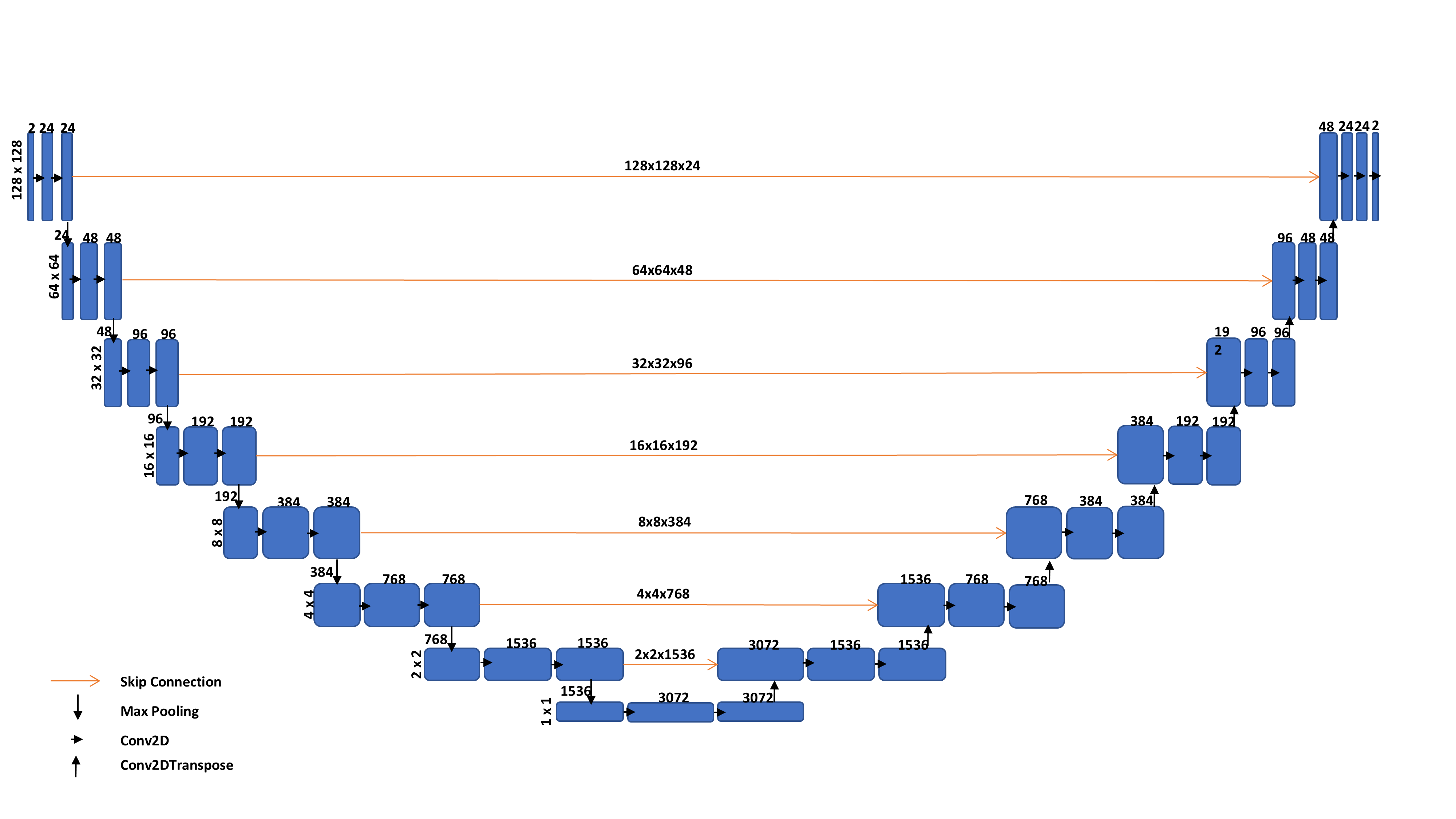}
\end{center}
\caption{\textcolor{blue}{The schematic representation of the U-Net architecture used in the Image Registration scheme.}}
\label{fig:Unet}
\end{figure}

The U-Net part of our image registration framework has a batch size of 128, unpadded, kernel size of the encoder is \textit{7x7}, the depth of the encoder is 1, and the filter size goes from 24 to 3072 in multiples of 2 for every step in the encoder. Batch normalization is used to allow the hidden layers to adjust effectively to the new distribution by avoiding the internal covariance shift that occurs due to the change in input distribution to the model \cite{ioffe2015batch}. The decoder has a kernel size \textit{3x3} and the skip connection is used to make up for the loss of information in the expansive path from the contracting path. The U-Net in Figure \ref{fig:Unet} is followed by the Spatial Transformation Network (STN) \cite{jaderberg2015spatial}, and the sampler within the STN produces a dense displacement field of the same size as the input image. The source image is warped onto the target image using the displacement field by applying the bilinear interpolation. The model is trained with a learning rate of $1\text{e-}4$ for 2000 epochs, and the Adam optimizer was used \cite{Kingma2015AdamAM}. The loss function is a combination of Cross Correlation (CC) and a regularizer to generate smooth displacement fields:
%
    \begin{equation}
	  \begin{array}{ll}
	   L_{\text{CC}} & = \operatorname*{arg\,min}_\theta \quad 1 - \frac{\sum_{i = 1}^{N} (q_i^* - \overline{q^*}) (q_i - \overline{q})}{\sqrt{(\sum_{i = 1}^{N} (q_i^* - \overline{q^*})}\sqrt{(\sum_{i = 1}^{N} (q_i - \overline{q})}} \\  
      L_{\text{smo}} & = \sum_{i = 1}^{N}||\boldsymbol{\nabla} {u'}|| \,,
	  \end{array}	
    \label{eq:losses}
    \end{equation}

where $q_i^*$ is the predicted source image on the $i$-th image pixel, $q_i$ is the original source image for the given network parameter $\theta$ and $N$ is the number of image pixels, $u'$ is the predicted displacement field. The total loss is a weighted sum of CC, and the regularizer term is given in Eq. \ref{eq:loss_tot}. In our experiment, we set the weights $\lambda _1$ and $\lambda _2$ as 1 and 0.001 respectively.

      \begin{equation}
        L_{\text{total}} = \lambda _1  L_{\text{CC}} + \lambda _2 L_{\text{smo}} 
      \label{eq:loss_tot}
      \end{equation}

\subsection{Estimation of the Poisson's ratio from the image displacement using reformulated Lam{\'e}-Navier operator}

The values of the Poisson's ratio on image pixels can formally be estimated directly from the Lam{\'e}-Navier PDE (Eq. \ref{eq:lame-navier-pde-geo}). By multiplying the vector-valued Eq. \ref{eq:lame-navier-pde-geo} by the vector
$\boldsymbol{\Delta}{u}$, one obtains a scalar equation which can be resolved with respect to $\nu$ resulting as follows:
\begin{equation}
    \nu = \frac{1}{2}\,\left(\frac{\boldsymbol{\Delta}{u} \cdot \text{grad} \, \text{div} \, {u}}{\boldsymbol{\Delta}{u} \cdot \boldsymbol{\Delta}{u}} + 1 \right)\,.
\label{eq:nu_local}
\end{equation} 

The identity Eq. \ref{eq:nu_local} defines the local value of the Poisson's ratio as a ratio of differential operators of the Lam{\'e}-Navier PDE. For a 128x128 image, Eq. \ref{eq:nu_local} yields a map of 128x128 Poisson's ratio values, where the value for every image pixel is computed using the numerical difference scheme for the first and second-order spatial derivatives.
%
%
In the case of correctly computed image displacement ($u$), for example as a numerical solution of a BVP for the given value of material compressibility $\nu$,  the average value of the 128x128 map of Poisson's ratios from Eq. \ref{eq:nu_local} is equal to the 'real value' of $\nu$. However, when reconstructed from image series, the displacement field is always overlaid with noise. Consequently, the Poisson’s ratio map obtained from Eq. \ref{eq:nu_local} becomes more dispersed. In the worst case, the average of such a noisy $\nu$ distribution can substantially deviate from the correct value of the Poisson's ratio. 

\subsection{Estimation of the Poisson's ratio from image displacement using deep learning}

To determine the compressibility of the material from the image displacement, a deep learning model based on Convolutional Neural Networks (CNN) was developed. This model takes a pair of images and effectively performs the regression (i.e. end-to-end mapping) of reconstructed displacement directly onto the single value of Poisson's ratio for the entire domain. This means that, unlike the above-described approach (Eq. \ref{eq:nu_local}), which determines local values of Poisson's ratio (and thus a distribution of $\nu$ for the entire image), CNN is trained to predict a single global value of $\nu$ of the entire image domain.
We adopted a data-driven approach and trained the CNN on ground truth FDM solutions to predict the value of Poisson's ratio directly from the reconstructed displacement field. The deep learning model is trained using a set of reference displacements computed using the FDM for different values of material compressibility. In particular, our framework relies on a CNN architecture, which has a series of intermediate hidden layers of filter size 16 to 1024 with a maxpooling of 2. The kernel size is maintained at 7, followed by a global convolution and a series of dense layers until the desired size of output 1. The model is trained with a batch size of 128, trained for 500 epochs with a learning rate of $1\text{e-}4$. The loss function used for the regression task is the mean squared error, given as follows:
      \begin{equation}
       L_{\text{MSE}} = \operatorname*{arg\,min}_\theta \quad((\nu;\theta) - \nu^* )\,. 
      \label{eq:loss_mse}
      \end{equation}
A customized 'Sigmoid' activation function is used in the last layer to restrict predicted values of Poisson's ratio within the physically admissible bounds of $\nu\in[0, 0.5]$:
       \begin{equation}
       \sigma_{\nu}(x) = 0.5\left(\frac{1}{1 + e^{-x}}\right)\,. 
      \label{eq:sigmaPR}
      \end{equation}


\subsection{Assessment and modelling of inaccuracies of the displacement field reconstructed from image registration}

The vector field $u'(x)$ mapping of source on target images, predicated by image registration, certainly deviates from the original displacement field $u(x)$ computed by the FDM.  
The difference between predicted and original displacements 
$n(x) = u'(x) - u(x)$ is further termed as 'noise', and the displacement overlaid with such 'noise' as 'noisy displacement'. 
%
%
For analysis of the effects of imprecise displacement reconstruction using image registration, a general assessment of the noise properties is required. For this purpose, a statistical map of the dependencies between the angle of the noisy displacement and the magnitude of the local displacement were assessed. Thereby, the angle $\theta$ between predicted and reference displacement was computed using the MATLAB function $vrrotvec$, and the magnitude of the displacement $||u||$ using the function $norm$. 
%
The rationale behind the assessment of the relationship $\theta(||u||)$ is that larger errors of image registration are expected for regions with smaller displacement magnitude, i.e, where the registration algorithm lacks structural differences between source and target images to reconstruct correct displacement vectors. The relationship $\theta(||u||)$ was computationally established as a distribution of bins, splitting the magnitude $||u||$ equally into 10 bins and collecting the $\theta$ corresponding to the magnitude into the bins. To systematically study the effects of noise on the Poisson's ratio computed using Eq. \ref{eq:nu_local}, a series of 'noisy displacement' was computed as follows
%
%
       \begin{equation}
       \label{eq:iterNoise}
        {u_{n}}\left(\phi(x)\right) = u\left(\phi(x) \pm \alpha \, \theta(||u(x)||)\,\right)\,,
      \end{equation}
where $u(\phi(x))$ is the reference displacement with a certain spacial orientation $\phi(x)$ computed using FDM,  ${u_{n}}(\phi(x))$ is the noisy displacement calculated from $u(\phi(x))$ under consideration of the local dependency of the deviation angle $\theta$ from the displacement magnitude $||u(x)||$, and $\alpha\in[0,1]$ is the scaling factor of the noise level.

\subsection{Training of deep learning based non-rigid image registration and Poisson's ratio estimation models}
The training of both the non-rigid image registration and regression models was implemented using TensorFlow 2.047 with the Keras API and Python 3.8. The libraries Numpy \cite{walt2011numpy}, Scikit-image \cite{van2014scikit}, and OpenCV \cite{opencv_library} were used for other utility operations. The models were trained on NVIDIA A100 SXM4 GPU featuring 80 GB RAM, running on a Linux operating system.

\subsection{Validation of deep learning based non-rigid image registration and Poisson's ratio estimation models}
The predictions of both the deep learning-based non-rigid image registration were validated using the root mean squared error (RMSE) and standard deviation (SD) as follows
       \begin{equation}
       \label{eq:maeMag}
       \begin{gathered}
        RMSE = \sqrt{\frac{1}{N}\sum_{i=1}^{N}\left(u_i - \hat{u}_i\right)^2}
        \\
        SD = \sqrt{\frac{ \sum_{i = 1}^{N} (u^{(i)} - \overline{u})^2 }{N}}\,.
      \end{gathered}
      \end{equation}
      
The predictions of both the deep learning-based Poisson's ratio estimation models were validated using the mean and standard deviation (SD) and compared against the ground truth Poisson's ratio.

\section{Results}
\begin{figure}[ht!]
\begin{center}
\includegraphics[width=18cm]{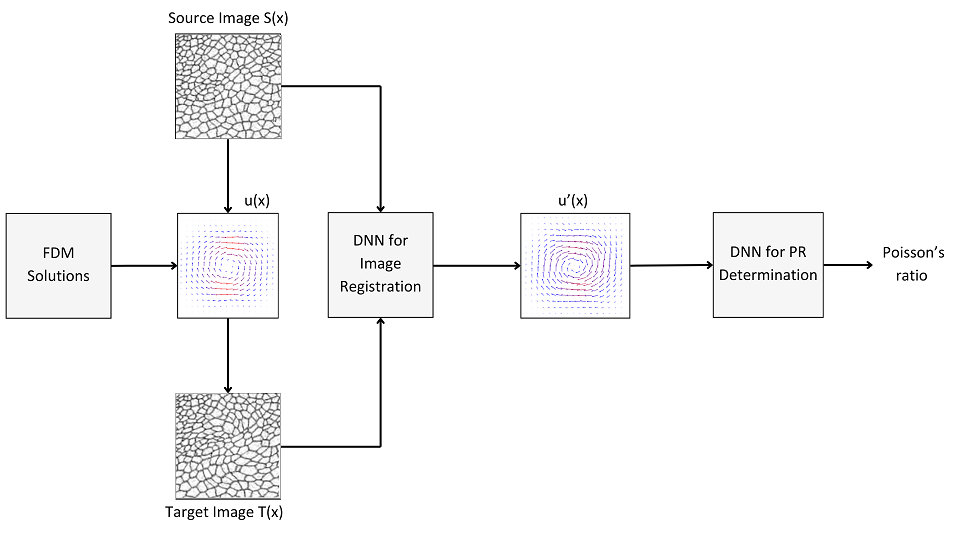}
\end{center}
\caption{Overview of the data flow and the computational framework for estimation of Poisson's ratio from image series (from left to right): FDM is used to compute
the reference displacement $u(x)$ for different boundary value problems and values of Poisson's ratio; source image is warped using the FDM-computed displacements to generate target image $T(x)=S(x+u)$; image registration is used to recover the displacement field $u'(x)$ from a pair of source and target (i.e. warped source) images; Poisson's ratio of unseen image pairs is predicted from the displacement $u'(x)$ using a DNN model which was trained on the reference set of displacements and corresponding values of Poisson's ratio.}
\label{fig:schematics}
\end{figure}

Our study results include the following major steps: (i) generation of non-rigid image displacements (${u}$) using FDM solutions of randomly defined linear elastic BVPs, (ii) Reconstruction of the displacement field from pairs of original and target images using image registration and, (iii) estimation of the Poisson's ratio from the reconstructed displacement field using a pre-trained CNN model. Figure \ref{fig:schematics} shows the visual representation of our framework to determine the material properties.

\subsection{Generation of reference displacement fields corresponding to predefined (i.e, known) values of material compressibilities}
\begin{figure}[ht!]
\begin{center}
\includegraphics[width=16cm]{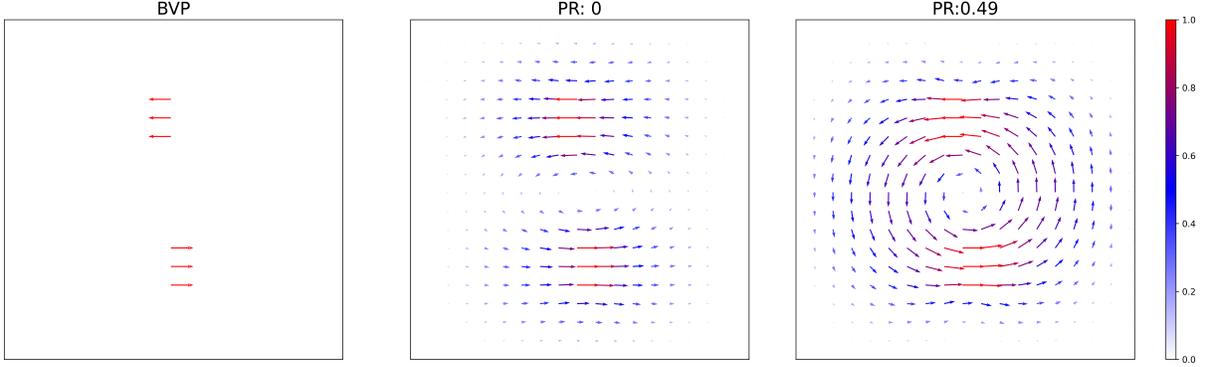}
\end{center}
\caption{An example of a linear elastic BVP and corresponding FDM solutions for two distinctive values of Poisson's ratio $\nu=0$  and $\nu=0.49$ corresponding to high and low compressible material approximations, respectively: (left) a BVP with prescribed boundary displacements, (middle) FDM solution for the Poisson's ratio $\nu=0$ (high compressible material model), (right) FDM solution for the Poisson's ratio $\nu=0.49$ (low compressible material model). For better visualization, displacement vectors are shown for every fifth pixel of the 128x128 image.}
\label{fig:fdmsoln}
\end{figure}

For the simulation of linear elastic materials with different compressibilities, a set of 4854 randomly defined boundary value problems and corresponding reference solutions (i.e, whole image displacement fields) was generated using FDM as described in the Methods section. Thereby, three distinctive values of Poisson's ratio ($\nu = 0, 0.25, 0.49$) were used for the simulation of high, intermediate, and low compressibility, respectively, for different boundary value problems. As a result, a set of 4854 whole image displacement fields as a function of the Poisson's ratio (${u}_i(\nu_i)$) was computed, where $i=[1,4854]$ is the index of each simulation based on a randomly defined boundary value problem and a randomly selected value of the Poisson's ratio out of the three distinctive values.
Figure \ref{fig:fdmsoln} shows examples of two FDM solutions corresponding to $\nu = 0$ and $\nu = 0.49$ for a simple BVP given by the displacement of the central image pixel. As one can see, patterns of displacement fields for low and high-compressible materials exhibit distinctive differences. In particular, the low compressible medium exhibits vortices due to volume preservation that are not present in the deformation pattern of the high compressible material.

\subsection{Generation of reference pairs of source and non-rigidly displaced target images}
\begin{figure}[ht!]
\begin{center}
\includegraphics[width=14cm]{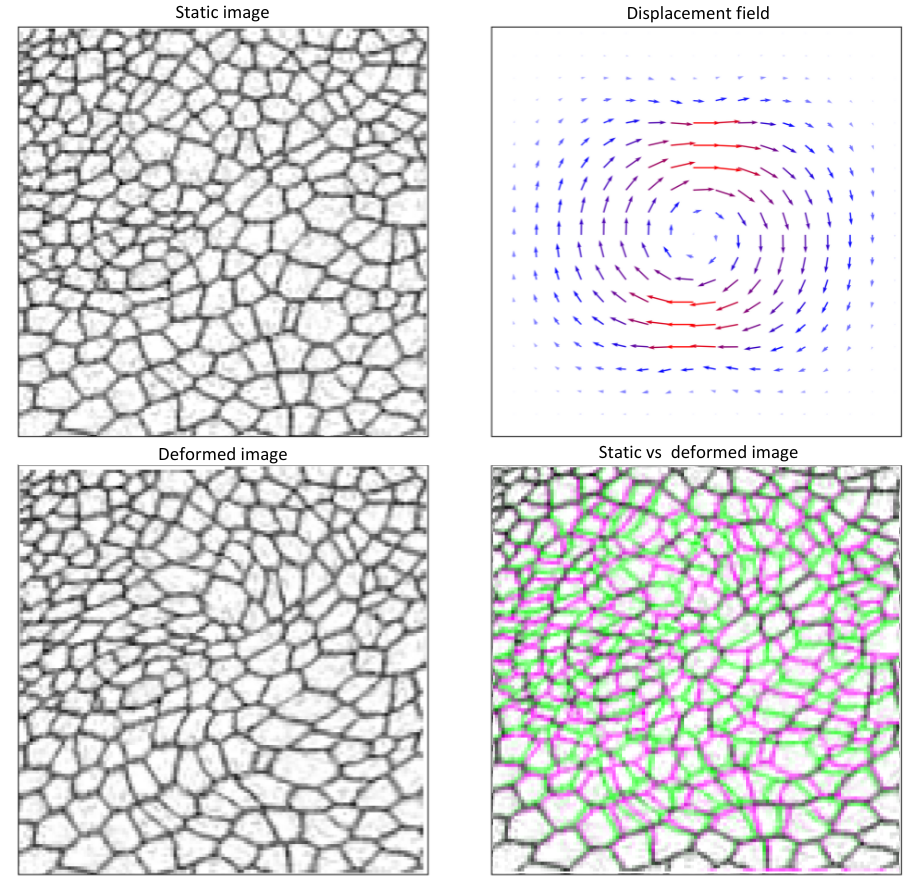}
\end{center}
\caption{An example of image warping using displacements computed as a FDM solution of linear elastic BVPs: (top, left) source (static) image ($S(x)$), (top, right) FDM-computed displacement field $u$, (bottom, left) target image computed as deformed version of the source images using the FDM displacement $T(x)=S(x+u)$, (bottom, right) overlay of source and target images.}
\label{fig:stat_mov}
\end{figure}

To simulate a realistic example of the biologically relevant problem, the reference images in this study were derived from a time series that depicts embryonic development of Drosophila wing tissue \cite{10.7554/eLife.07090}. 
An example of a pair of 128x128 static source and deformed target images is shown in Figure \ref{fig:stat_mov}. Pairs of source and target images generated for different displacements resulting from different linear elastic BVP solutions were subsequently used for the reconstruction of displacement ${u'}$ using image registration. The advantage of our synthetic-data approach is that the reconstructed displacement ${u'}$ can be compared to the original displacement ${u}$ computed as a solution of a linear elastic BVP with a particular value of Poisson's ratio. Consequently, this allows quantitative assessment of the accuracy of the displacement reconstructed using image registration vs. the exact displacement of the elastic continuum with a particular (known) material compressibility.

%

\subsection{Automated non-rigid image registration using a deep learning model}

\begin{figure}[ht!]
\begin{center}
\includegraphics[width=14cm]{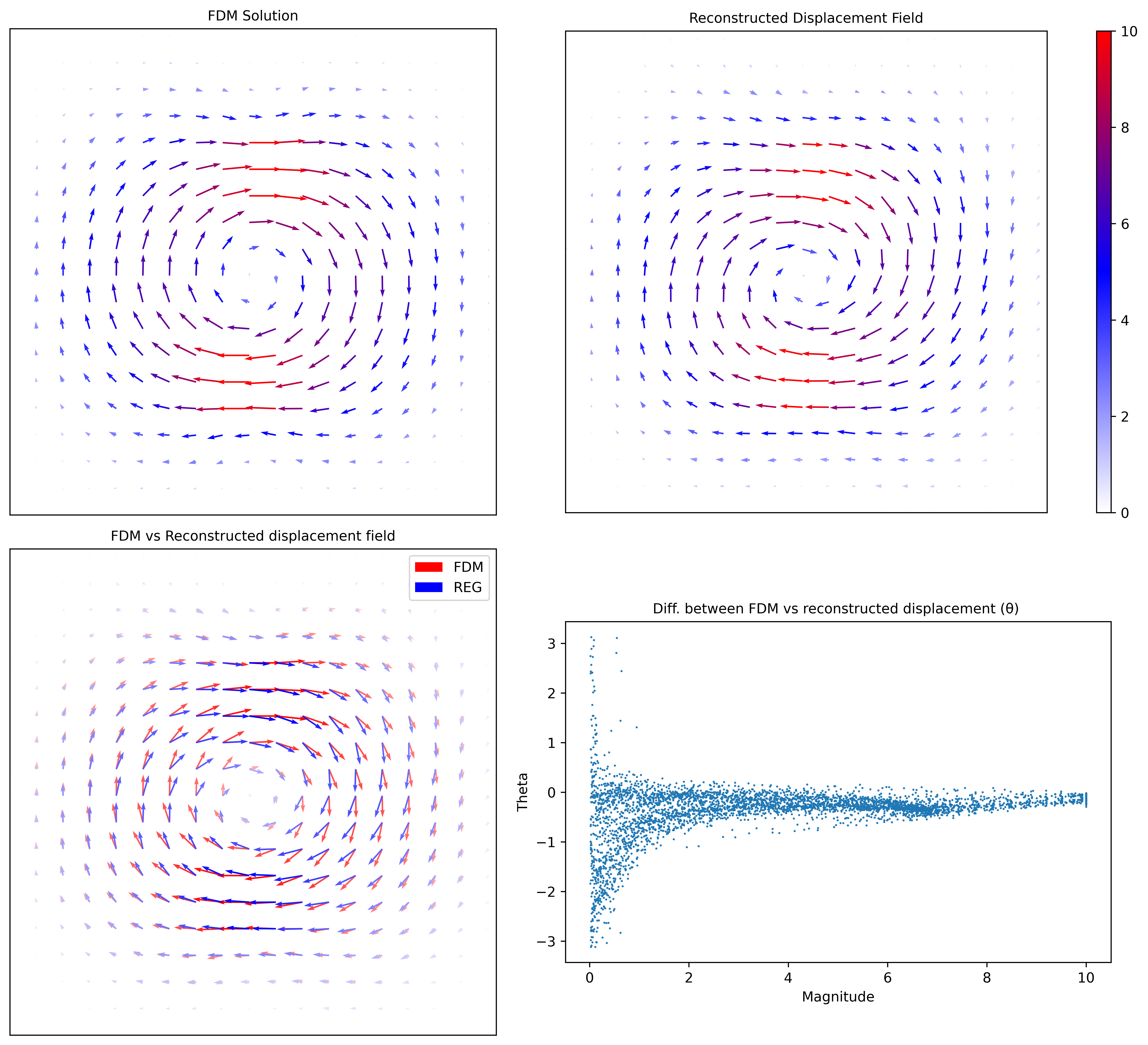}
\end{center}
\caption{Comparison of image displacements computed using FDM vs. image registration: (top, left): displacement field computed as an FDM solution of linear elastic BVP, (top, right) displacement reconstructed using deep learning based image registration for a pair of source and target images, (bottom, left) overlay of displacements obtained from FDM (red arrows) vs. displacement reconstructed using image registration (blue arrows), (bottom, right) visualization of the angle between FDM-computed and DNN-reconstructed displacements as a function of displacement magnitude.}
\label{fig:image_reg_disp}
\end{figure}

A deep learning model for non-rigid image registration was applied to predict the displacement field from a pair of source (fixed) vs. target (deformed) images. To assess the accuracy of non-rigid image registration by prediction of the image displacement, a comparison of original and reconstructed displacement fields was performed. Table \ref{tab:perfMetfwd} summarizes the performance of image registration in terms of mean/standard deviation of differences between the magnitude of reference FDM vs. reconstructed displacement fields, and Figure S1 (Left) in supplementary information shows the RMSE between reference FDM and the reconstructed displacement as a function of Poisson's ratio. As one can see, the displacement field reconstructed from image registration shows a large similarity with the original reference displacement field. To stick to the linear elastic bounds of material deformation, the maximum displacement of 128x128 images was kept up to 10 pixels for all BVPs computed in this study. An example and comparison of differences between the exact FDM solution and displacement reconstructed from image registration is shown in Figure \ref{fig:image_reg_disp}.

\begin{table}[]
\centering
\caption{\label{tab:perfMetfwd}Summary of performance of deep learning model by the reconstruction of image displacement from all 4584 image pairs vs. reference displacement computed using FDM for three distinctive values of Poisson's ratio.}
\begin{tabular}{|c|c|c|c|}
\hline
Metrics & $\nu=0$ & $\nu=0.25$ & $\nu=0.49$ \\ \hline
RMSE     & 0.88 & 0.88    & 1.0    \\ \hline
SD      & 0.22 & 0.18    & 0.50    \\ \hline
\end{tabular}

\end{table}
%
%

%
%
%
%

%
\subsection{Estimation of the Poisson's ratio from the displacement field using the Lam{\'e}-Navier PDE}

\begin{figure}[ht!]
\begin{center}
\includegraphics[width=14cm]{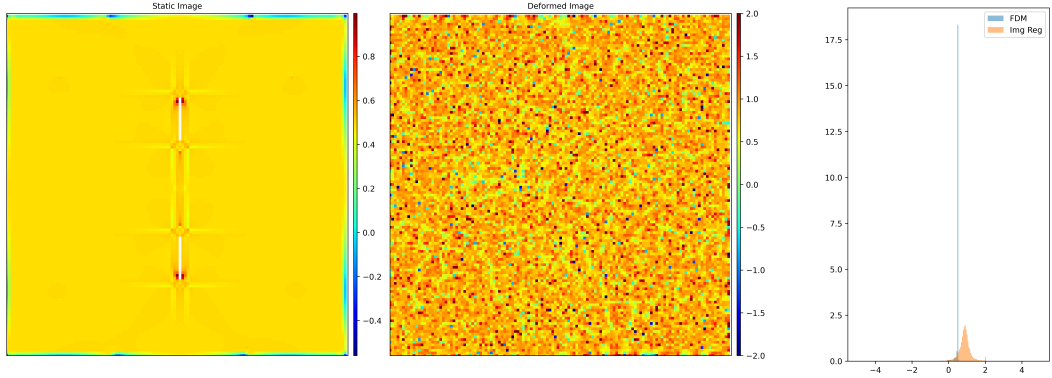}
\end{center}
\caption{Comparison of the Poisson's ratio maps reconstructed from the reformulation of Lam{\'e}-Navier PDE with respect to local $\nu$ values (see Eq. \ref{eq:nu_local}): (left) $\nu$ map corresponding to the exact FDM solution (displacement field), (middle) $\nu$ map corresponding to the displacement reconstructed from image registration, (right) histograms of $\nu$ distributions for all image pixels computed from FDM re. image registration based displacement fields. In this example, the correct value of Poisson's ratio is $\nu=0.49$.}
\label{fig:nu_dis}
\end{figure}

The values of the Poisson's ratio are determined from the displacement field using a reformulation of the Lam{\'e}-Navier PDE, see Eq. \ref{eq:nu_local}. In particular, meaningful values of the Poisson's ratio are determined for the image regions (pixels) with the so-called free boundary conditions.
Within the regions with Dirichlet boundary conditions, the displacement does not necessarily fulfil the Lam{\'e}-Navier PDE and, thus, cannot be expected to lead to meaningful values of the Poisson's ratio using  Eq. \ref{eq:nu_local}.
Figure \ref{fig:nu_dis} shows examples of $\nu$-maps calculated for the FDM-computed (left) vs. reconstructed image displacement (middle, right). 
As one can see, the average value of the $\nu$ distribution for the reference displacement (i.e. FDM solution) computed using Eq. \ref{eq:nu_local} corresponds to the exact $\nu$ value (in this example, $\nu=0.49$). In contrast, the distribution of Poisson's ratio for the displacement reconstructed from the registration of source and target images is substantially deviating from the exact value. Figure \ref{fig:image_reg_disp} (top, right) shows that the displacement estimated from the image registration resembles the reference (FDM-computed) displacement for $\nu=0.49$. 
However, the Poisson's ratio cannot be reconstructed from this slightly deviating displacement field and corresponds to the correct value only for a minor fraction ($10^{-3}$) of all image pixels. This shows that despite an overall similarity between exact FDM-computed and registration-estimated displacement fields, the Poisson's ratio cannot be accurately determined as a quotient of differential operators from Eq. \ref{eq:nu_local}, because they are highly sensitive to subtle changes in the relative orientation and magnitude of neighbouring displacements that could not be reconstructed accurately enough by the non-rigid image registration.
To determine the critical level of noise for the reconstruction of Poisson's ratio from the image displacement, a series of reference (FDM-computed) displacements overlaid with a
progressively larger level ($\alpha>0$) of additive noise were generated using Eq. \ref{eq:iterNoise}.
It turned out that the accuracy of Poisson's ratio was calculated using Eq. \ref{eq:nu_local} dramatically drops beginning from a relatively low level of noise: $\alpha>0.006$ which is 0.6\% of the modeled noise, presented in Figure S2.

\subsection{Estimation of the Poisson's ratio from the displacement field using the deep learning approach}

\begin{figure}[ht!]
\begin{center}
\includegraphics[width=16cm]{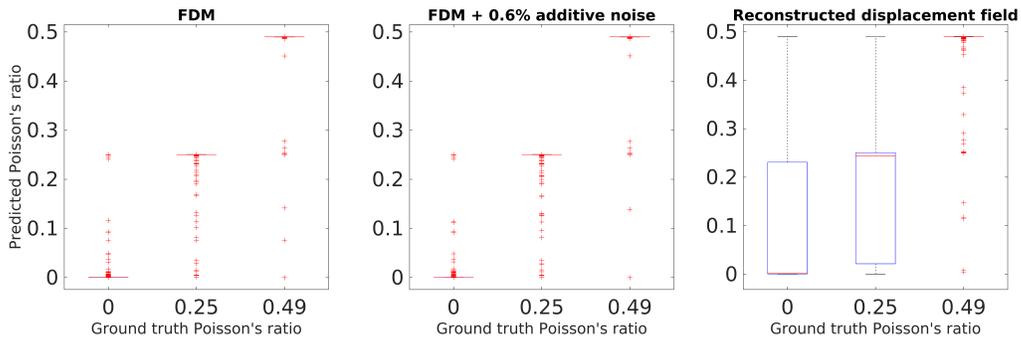}
\end{center}
\caption{Boxplot visualization of the Poisson's ratio distributions obtained for all test images re. displacement fields trained using exact FDM solutions without noise: (left) displacement obtained as an exact FDM solution, (middle)  FDM-computed displacement overlaid with 0.6\% additive noise and  (right) displacement fields reconstructed from image registration.}
\label{fig:box_plot}
\end{figure}

The estimation of Poisson's ratio using the end-to-end deep learning approach was carried out in a fully data-driven manner. That is, the deep learning model was trained on a set of displacement fields (solutions of BVPs) as input and the corresponding values of Poisson's ratio as output. In total, three models were trained on displacement fields computed as exact FDM solution as well as FDM solution overlaid with 5\% and 10\% noise, respectively. The models were validated with three different types of dataset: 
(i) unseen FDM solutions whose patterns are similar to the ground truth FDM solutions used to train the model, and  (ii) FDM solutions overlaid with additive noise (referred to as $u_{n}$ in the Method section). The third dataset used to validate the deep learning model was (iii) the displacement reconstructed from the non-rigid image registration. Our experimental results show that the deep learning model performs accurately with the exact FDM solutions and the FDM solutions overlaid with 0.6\% additive noise ($u_{n}$). 
 In the case of the displacement field reconstructed from the image registration, the accuracy of $\nu$ predictions becomes less accurate, but in a $\nu$-dependent manner. The models trained on FDM, FDM + 5\% noise and FDM + 10\% are comparable to each other.
Table \ref{tab:prtab_box} summarizes the performance of our deep learning model by prediction of three distinctive values of Poisson's ratio vs. conventional approach based on the reformulation of Lam{\'e}-Navier PDE (Eq. \ref{eq:nu_local}). The relationship between RMSE of reference FDM and reconstructed displacement field as a function of Poisson’s ratio with the difference in predicted and ground truth Poisson’s ratio using the model trained on reference FDM validated against the reconstructed displacement field as a function of Poisson's ratio is presented in Supplementary material Figure S1. Performance measures (Mean/SD) were assessed for all 4584 'unseen' displacements from the validation dataset. As one can see, DNN shows superior performance compared to the conventional approach for all values of Poisson's ratio and all data modalities, including exact FDM displacement, FDM displacement overlaid with additive noise, and even more noisy displacement obtained from image registration.
  Figure \ref{fig:box_plot} shows boxplot visualization of DNN predicted values of Poisson's ratio for all tested image pairs vs. three ground truth values. The results for DNN predicted values of Poisson's ratio predicted using models trained on FDM + 5\% noise and FDM + 10\% noise are provided in the supplementary material, Figure S3 and Figure S4. DNN predicts Poisson's ratio more accurately on displacement fields corresponding to incompressible ($\nu=0.49$) compared to more compressible materials ($\nu=0$ and $\nu=0.25$).
\begin{table}[]
\centering
\caption{\label{tab:prtab_box}{Comparison of performance of conventional approach (Eq. \ref{eq:nu_local}) and DNN  by prediction of three distinctive values of Poisson's ratio on the whole set 'unseen' validation displacements.}}
\begin{tabular}{|c|c|cc|cc|cc|}
\hline
\multirow{2}{*}{Method} & \multirow{2}{*}{Displacement} & \multicolumn{2}{c|}{$\nu=0$} & \multicolumn{2}{c|}{$\nu=0.25$} & \multicolumn{2}{c|}{$\nu=0.49$} \\ \cline{3-8} 
 &  & \multicolumn{1}{c|}{Mean} & SD & \multicolumn{1}{c|}{Mean} & SD & \multicolumn{1}{c|}{Mean} & SD \\ \hline
\multirow{3}{*}{\begin{tabular}[c]{@{}c@{}}Reformulation of \\ Lam{\'e}-Navier PDE \\ w.r.t. local $\nu$ (Eq. \ref{eq:nu_local})\end{tabular}} & FDM & \multicolumn{1}{c|}{-0.01} & 0.01 & \multicolumn{1}{c|}{0.24} & 0.01 & \multicolumn{1}{c|}{0.48} & 0.00 \\ \cline{2-8} 
 & FDM + 0.6\% noise & \multicolumn{1}{c|}{0.47} & 0.08 & \multicolumn{1}{c|}{0.48} & 0.06 & \multicolumn{1}{c|}{0.54} & 0.05 \\ \cline{2-8} 
 & Image reg. & \multicolumn{1}{c|}{0.88} & 0.03 & \multicolumn{1}{c|}{0.88} & 0.03 & \multicolumn{1}{c|}{0.84} & 0.04 \\ \hline
\multirow{3}{*}{DNN model of global $\nu$} & FDM & \multicolumn{1}{c|}{0.00} & 0.02 & \multicolumn{1}{c|}{0.25} & 0.03 & \multicolumn{1}{c|}{0.49} & 0.03 \\ \cline{2-8} 
 & FDM + 0.6\% noise & \multicolumn{1}{c|}{0.00} & 0.02 & \multicolumn{1}{c|}{0.25} & 0.02 & \multicolumn{1}{c|}{0.49} & 0.02 \\ \cline{2-8} 
 & Image reg. & \multicolumn{1}{c|}{0.08} & 0.12 & \multicolumn{1}{c|}{0.18} & 0.13 & \multicolumn{1}{c|}{0.48} & 0.04 \\ \hline
\multirow{3}{*}{\begin{tabular}[c]{@{}c@{}}DNN model of global $\nu$\\ Trained on\\ FDM + 5\% noise\end{tabular}} & FDM & \multicolumn{1}{c|}{0.00} & 0.01 & \multicolumn{1}{c|}{0.25} & 0.02 & \multicolumn{1}{c|}{0.49} & 0.03 \\ \cline{2-8} 
 & FDM + 0.6\% noise & \multicolumn{1}{c|}{0.00} & 0.01 & \multicolumn{1}{c|}{0.25} & 0.01 & \multicolumn{1}{c|}{0.49} & 0.02 \\ \cline{2-8} 
 & Image reg. & \multicolumn{1}{c|}{0.07} & 0.10 & \multicolumn{1}{c|}{0.20} & 0.09 & \multicolumn{1}{c|}{0.48} & 0.05 \\ \hline
\multirow{3}{*}{\begin{tabular}[c]{@{}c@{}}DNN model of global $\nu$\\ Trained on\\ FDM + 10\% noise\end{tabular}} & FDM & \multicolumn{1}{c|}{0.00} & 0.03 & \multicolumn{1}{c|}{0.24} & 0.03 & \multicolumn{1}{c|}{0.49} & 0.02 \\ \cline{2-8} 
 & FDM + 0.6\% noise & \multicolumn{1}{c|}{0.00} & 0.02 & \multicolumn{1}{c|}{0.24} & 0.03 & \multicolumn{1}{c|}{0.49} & 0.02 \\ \cline{2-8} 
 & Image reg. & \multicolumn{1}{c|}{0.10} & 0.12 & \multicolumn{1}{c|}{0.20} & 0.12 & \multicolumn{1}{c|}{0.48} & 0.04 \\ \hline
\end{tabular}
\end{table}
 
%





\subsection{Grad-CAM analysis of model predictions}
\begin{figure}[ht!]
\begin{center}
\includegraphics[width=14cm]{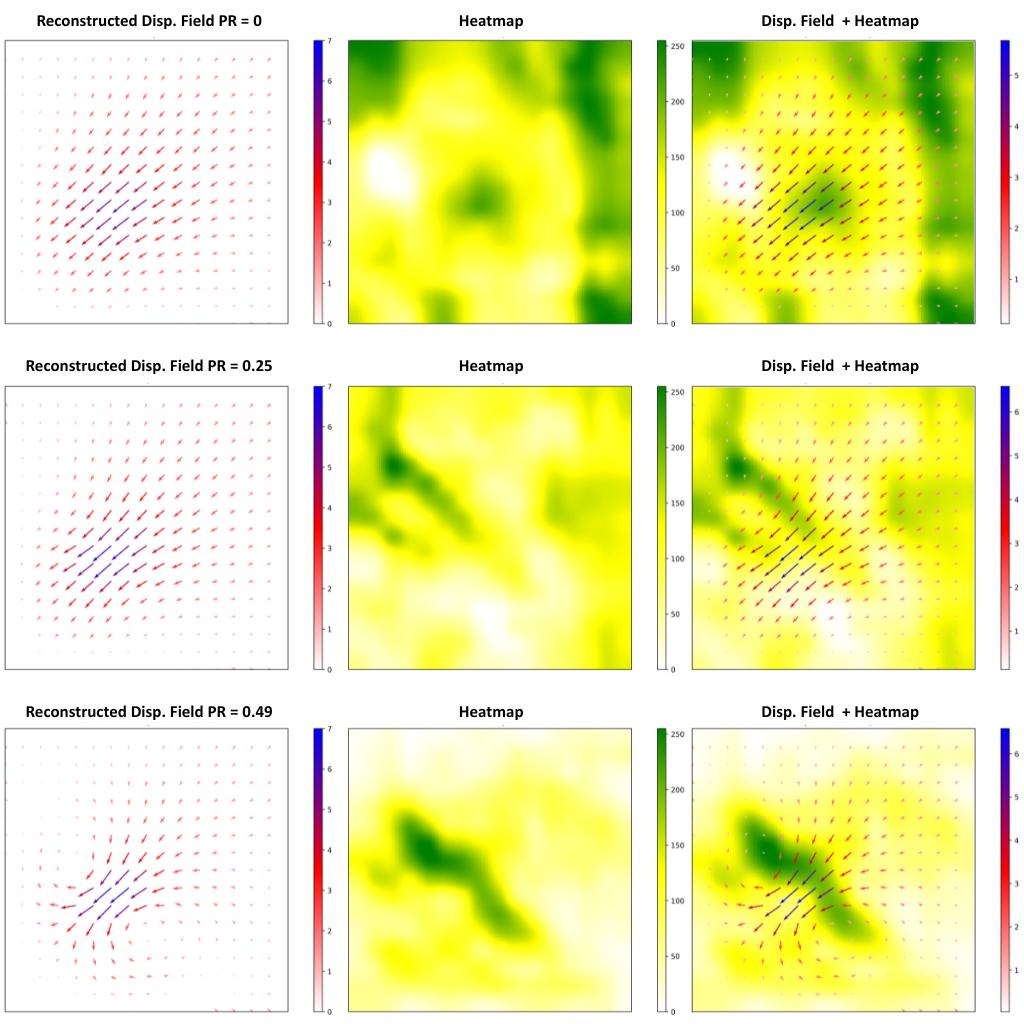}
\end{center}
\caption{Examples of Grad-CAM analysis of intermediate network layers corresponding to correct prediction of three different Poisson's ratio values ($\nu=[0~0.25~0.49]$) from the whole image displacement field. As one can see, the image regions relevant to model decision making correspond to areas with higher probability of occurance of vortices, namely in the vicinity of the large prescribed displacement along the line perpendicular to the displacement direction, - similar to paddle eddies. In the case of more compressible materials ($\nu=[0~0.25]$), the activation patterns also turn out to be diffuse compared to incompressible displacement field ($\nu=0.49$).}
\label{fig:grad_cam}
\end{figure}

In order to qualitatively assess displacement features that mostly contributed to the model prediction, Grad-CAM analysis was performed \cite{selvaraju2017grad}. The Gradient-weighted Class Activation Mapping (Grad-CAM) enables qualitative assessment of contributions from CNN layers to the final result of model predictions in terms of heatmaps that highlight input regions most influential to the model's decision. Figure \ref{fig:grad_cam} shows three examples of Grad-CAM heatmaps corresponding to predictions of three different values of the Poisson's ratio. From these three but also further examples, one can see that (i) activation patterns of intermediate model layers appear to be more compact for displacements of incompressible materials ($\nu=0.49$) and, in contrast, more diffuse for displacement of compressible materials ($\nu=0$ and 
$\nu=0.25$); (ii) furthermore, the model seems to highlight the 'paddle-eddies' like regions where displacement fields (may) exhibit strong vortices, especially in the case of incompressible material behavior, see. Figure \ref{fig:grad_cam}.


\section{Discussion and Conclusion}

In this work, we have presented an end-to-end deep learning framework for automated contactless estimation of material compressibility from a series of source-target images exhibiting a relative elastic deformation. 
The linear elastic Finite Difference Method was used for the generation of reference displacement fields.
Our method is based on the sequential application of two deep neural networks: (i) one for the estimation of the displacement field using the non-rigid image registration, and (ii) another one determining the material compressibility of the whole image domain as a result of end-to-end regression of the reconstructed displacement field onto the single value of Poisson's ratio. 
%
%
A modification of the state-of-the-art VoxelMorph image registration was used in this work for the reconstruction of image displacement.
However, the displacement field reconstructed from image registration always contains a certain amount of error. 
Here, we observed that even a very small deviation of reconstructed displacement from the FDM reference severely affects the accuracy of compressibility predictions using the conventional PDE-based approach.  
%
%
%
%
%
In contrast to local differential quotients of the Lam{\'e}-Navier PDE, our end-to-end deep learning model was capable of determining Poisson's ratio from imperfectly reconstructed, noisy displacement more accurately. In particular, deep learning predictions turned out to be more accurate for displacement fields from low-compressible material approximation (i.e, $\nu=0.49$).
This remarkable result demonstrates higher-order cognitive abilities of DNN that appear to rely on its capabilities to detect higher-order structural properties of displacement fields (i.e, occurrence of vortices) as indicated by our Grad-CAM analysis of network activation layers. 
In this feasibility study, we exemplarily focus on the estimation of the Poisson's ratio from the whole image displacement. The extension of this framework to a more complex material distribution is, in general, possible by adopting image-to-image approach instead of our image-to-number regression model. On the other hand, a single-value regression model can also be used for estimation of compressibility distributions when applied as a  sliding window approach.   
%
%
%
%
Our framework was evaluated on images warped by the linear elastic displacements that were computed for predefined values of material compressibility using FDM.
%
%
However, it can principally be applied for the estimation of Poisson's ratios of real materials. In this case, cross-validation against real physical measurements such as atomic force microscopy (AFM) or ultrasound elastography would be of advantage.
%
%
To reduce the number of time-consuming computations using our data-driven approach, training, and validation of deep learning models was carried out using images of relatively small spatial resolution of 128x128, which limits the model's applicability to high resolution images directly. However, the high resolution images can be processed as 128x128 masks.  
%
%
For analysis of larger images, supervised models have to be trained on large reference data or, alternatively, an unsupervised approach incorporating PDE-laws as a training criterion can be considered instead. 
Overall, our study demonstrates the great potential of deep learning models by remote sensing of physical properties from optical observation of material behavior. Incorporating the physical governing equations in the image registration scheme can eliminate the need for secondary DNN required for determining the Poisson's ratio which, requires special investigation and would be the future extension of the work.
%
\section*{Acknowledgements}


This work was supported by  the German Federal Ministry of Education and Research (BMBF) within the scope of the AVATARS project (FKZ 031B0770A).



\section*{Availability of Data and Materials}

Further datasets used and analyzed during the current study are available from the corresponding author upon reasonable request.

\bibliographystyle{unsrt}  
\bibliography{references}  

\end{document}